\documentclass[lettersize,journal]{IEEEtran}
\usepackage{amsmath,amsfonts}
\usepackage{algorithmic}
\usepackage{algorithm}
\usepackage{array}
\usepackage[caption=false,font=normalsize,labelfont=sf,textfont=sf]{subfig}
\usepackage{textcomp}
\usepackage{stfloats}
\usepackage{url}
\usepackage{verbatim}
\usepackage{graphicx}
\usepackage{cite}
\usepackage{graphicx}
\usepackage{verbatim}
\usepackage{amsmath}
\usepackage{amssymb}
\usepackage{ctable}
\usepackage{multirow}
\usepackage{url}
\usepackage{marvosym}
\usepackage{rotating}
\usepackage{color}
\usepackage{sidecap}
\usepackage{colortbl}
\usepackage{xcolor} 
\usepackage{hyperref}

\definecolor{grey}{HTML}{B2BEB5} 
\definecolor{mygreen}{HTML}{008000} 
\begin{document}
\sloppy 

\title{Contrastive Learning of Visual-Semantic Embeddings}

\author{Anurag Jain, and Yashaswi Verma 
\thanks{A. Jain is with Cadence Design Systems (India) Private Limited, Pune, India - 411006. 
Email: anuragja@cadence.com}
\thanks{Y. Verma is with the 
Department of Computer Science and Engineering, 
Indian Institute of Technology, Jodhpur, India - 342037. 
Email: yashaswi@iitj.ac.in}
}

\markboth{Journal of \LaTeX\ Class Files,~Vol.~14, No.~8, August~2021}%
{Shell \MakeLowercase{\textit{et al.}}: A Sample Article Using IEEEtran.cls for IEEE Journals}

\IEEEpubid{0000--0000/00\$00.00~\copyright~2021 IEEE}

\maketitle

\begin{abstract}
Contrastive learning is a powerful technique to learn representations that are semantically distinctive and geometrically invariant. While most of the earlier approaches have demonstrated its effectiveness on single-modality learning tasks such as image classification, recently there have been a few attempts towards extending this idea to multi-modal data. In this paper, we propose two loss functions based on normalized cross-entropy to perform the task of learning joint visual-semantic embedding using batch contrastive training. In a batch, for a given anchor point from one modality, we consider its negatives only from another modality, and define our first contrastive loss based on expected violations incurred by all the negatives. Next, we update this loss and define the second contrastive loss based on the violation incurred only by the hardest negative. We compare our results with existing visual-semantic embedding methods on cross-modal image-to-text and text-to-image retrieval tasks using the MS-COCO and Flickr30K datasets, where we outperform the state-of-the-art on the MS-COCO dataset and achieve comparable results on the Flickr30K dataset. 

\end{abstract}



\section{Introduction and Background}
\label{sec:intro}

\IEEEPARstart{J}{oint} embedding learning 
is a well-studied problem, 
where the objective is to learn 
a common embedding 
space for two (or more) diverse 
domains/modalities 
({\it e.g.} images and text) 
that represents their underlying 
structure and semantics. 
Such embeddings entail mapping of 
semantically similar 
samples from diverse modalities to 
similar locations, thus 
facilitating direct matching of 
samples from different modalities using 
simple vector operations such as 
dot product. 
As a result, these can enable a wide 
range of visual and language 
understanding tasks such as 
image/video tagging~\cite{wsabie,devise}, 
captioning~\cite{sbu,videocapnaacl} and 
retrieval~\cite{andrej,UVS}, visual question 
answering~\cite{askneurons15}, 
fine-grained recognition~\cite{cmfinegrained16}, 
zero-shot learning~\cite{devise}, 
bilingual word embeddings~\cite{bilingual13}, etc. 

In this paper, we focus on learnig visual-semantic embeddings for cross-modal 
image-text retrieval; {\it i.e.,} the 
retrieval of semantically relevant images given a caption, or of semantically relevant 
captions given a query image. 
Since it is not possible to match samples 
from diverse modalities represented in heterogeneous 
feature spaces directly, many of the 
existing approaches have 
addressed this problem by learning an embedding 
function that projects these features into a 
common subspace where these can be directly matched. 
Among the non-deep learning based techniques, 
one popular paradigm is to learn a common subspace 
such that the correlations among semantically similar 
samples are 
maximized~\cite{hardoon,rasiwasia10,gma,rasiwasia14,mvcca,ranjan15,scalablemlcca}. 
The deep learning based techniques approach this 
problem by learning non-linear projections 
for individual modalities 
that are subsequently fused to lead to a common subspace 
where heterogeneous samples can be directly 
matched~\cite{mRNN,embeddingNet,jointLearning,vsepp}. 
These non-linear projections are mostly learned jointly 
by imposing a (dis)similarity 
constraints on image and text features using cross-modal 
triplet 
constraints~\cite{vsepp,UVS,GCSFDIS,andrej,Zhu2015AligningBA}. 
These ranking-based constraints capture semantic correlations 
between modalities 
by optimizing the similarity 
between positive pairs to be more than that 
between negative pairs by a margin. 
While the 
primary objective of such constraint 
functions 
is to minimize semantic gap based on  similarity 
between cross-modal paired samples, their ability to learn 
a good projection gets hindered due to 
heterogeneity gap that is introduced by incompatible 
feature distributions in diverse modalities. 

To address this, in this paper, we propose to learn joint embedding 
space by integrating triplet ranking loss 
with 
contrastive representation 
learning~\cite{simclr,miem19,dataefficient,supcon}. 
Contrastive representation learning has 
been found to be an effective approach in reducing 
the heterogeneity gap between semantically similar 
samples. 
The underlying 
idea in these techniques is that given an anchor 
point, maximize its similarity with a ``positive'' 
sample and minimize its expected similarity with 
many ``negative'' samples in a mini-batch. 
Among these, 
the SimCLR~\cite{simclr} approach, 
which is also a representative state-of-the-art in 
contrastive representation learning for unimodal data, 
does this using 
the normalized temperature-scaled cross entropy (or NCE) 
loss. Because of its simplicity, SimCLR has 
been adapted for a 
variety of visual understanding 
tasks~\cite{afros20,misra20,ssmvn}. 
Recently, \cite{ssmvn} 
proposed an NCE-based 
loss to learn generalized similarity-preserving multi-modal 
representations using expected similarity between 
{\it all} negative 
multi-modal pairs; {\it i.e.}, for an anchor point 
from one modality, they consider negatives from all the 
modalities including those from the anchor's modality. 
However, in our preliminary experiments, we used this loss 
to learn a joint visual-semantic embedding using 
image-caption pairs, and found this 
to give not-so-good results on the cross-modal retrieval 
task. 
This is because in cross-modal retrieval 
tasks, we are interested in reducing the similarity 
between negative pairs {\em across} different modalities 
instead of negative pairs {\em within} the same modality. 
To incorporate this idea, our first contribution is a new 
NCE-based contrastive loss for learning joint embeddings 
that considers negatives from other 
modalities {\it excluding} the anchor's modality. 
We would like to emphasize that while this update may appear 
to be trivial, it provides a significant boost in empirical 
results as we will discuss later 
({\it c.f.} Table~\ref{tab:BaselineComparison}). 

Next, 
it was noted by~\cite{vsepp} that a triplet 
loss function that considers 
expected similarity with all the negatives 
may be biased and create problematic local minima, 
particularly in those problems where we have a structured 
output space 
such as learning joint visual-semantic embeddings, and 
advocated the use of only hard negatives. 
As our second contribution, 
we integrate this idea 
in the contrastive learning set-up, and propose 
second NCE-based contrastive loss for 
joint embedding learning. Specifically, 
we update our previous loss function by considering 
only the {\it hardest} negative from other 
modalities excluding the anchor's modality. 
We will show later that this update 
further improves the empirical results. 
As per our knowledge, this is the first attempt of 
incorporating the idea of using only the hardest negatives 
in NCE-based deep contrastive learning. 

To validate the effectiveness of our proposals, 
we extensively compare them with existing joint 
embedding learning approaches 
on cross-modal image-text retrieval tasks using 
the MS-COCO and Flickr30K datasets, where we 
outperform the state-of-the-art on the MS-COCO dataset and
achieve comparable results on the Flickr30K dataset. 

The paper is organized as follows. In 
Section~\ref{sec:Preliminaries}, we discuss the preliminaries 
on learning visual-semantic embeddings using hinge-based 
triplet ranking loss. 
Section~\ref{sec:OurApproach} describes
the proposed approach. 
In Section~\ref{sec:Experiments}, we 
present the experimental analyses, and 
finally Section~\ref{sec:Conclusion} presents the
conclusions and directions for future research.

\begin{figure*}[t] 
\centering
\begin{tabular}{c}
  {{\includegraphics[width=0.85\linewidth]{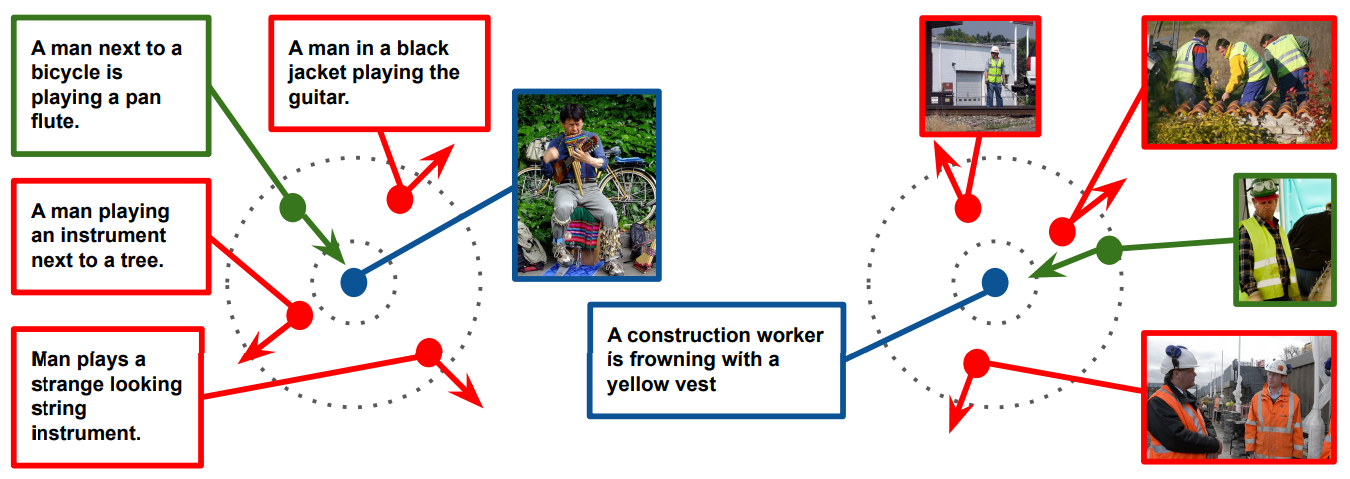}}}\\
  (a) \\
  {{\includegraphics[width=0.85\linewidth]{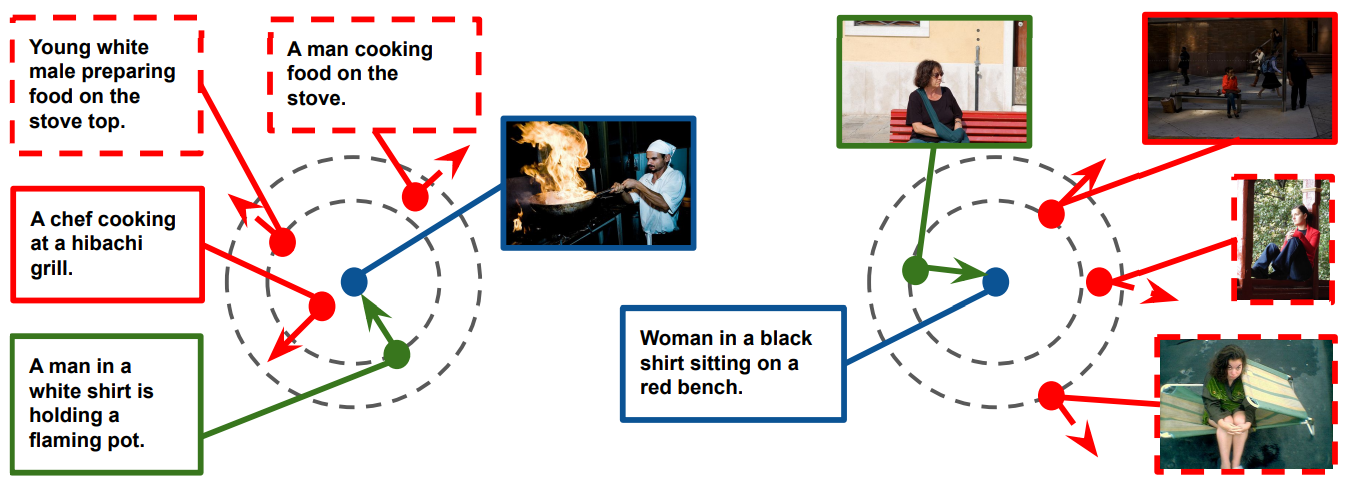}}}\\
  (b)
  \end{tabular}
\caption{A visual illustration of the 
proposed loss functions: (a) the sum-of-negatives 
loss $l_{CSN}$ (Eq.~\ref{eq:CMCL}), and (b) the 
max-of-negatives loss $l_{CMN}$ (Eq.~\ref{eq:cmnloss}). 
Data points marked in blue, green and red denote 
the anchor point, the positive (target) 
point and the negative points respectively 
(see text for details).}
\label{fig:illustration}
\end{figure*} 

\section{Preliminaries} 
\label{sec:Preliminaries}


To learn a visual-semantic embedding, 
our training set 
$\mathcal{D} = \{(I_i, C_i)\}$
consists of 
pairs of images and corresponding captions. 
We consider $(I_i, C_i)$ to denote a 
positive pair and $(I_i, C_j)$ ($j \neq i$) to denote a 
negative pair; {\it i.e.}, for image $I_i$,  $C_i$ is the most 
relevant caption and $C_j$ is a caption corresponding to some other image, and vice-versa. 
Initially, we use image and text encoders 
to compute the representations 
of an image and a caption respectively, and 
project them into a joint embedding space. 
For a given image $I$ and 
a caption $C$, we denote the embeddings in the joint embedding 
space as $z_{I}, z_{C} \in \mathbb{R}^d$ 
respectively. 
To compute the similarity between the embeddings of an image $I$ and a caption $C$ in this joint embedding space, we 
use normalized correlation (or, the cosine similarity), which 
is given by 
\begin{equation}
    s(I, C) = \frac{z_I^T z_C}{\|z_I\|  \|z_C\|}
    \label{eq:cosine}
\end{equation}



\subsection{Base Joint Embedding} 
\label{sec:BaseEmbedding} 


Many approaches that learn joint visual-semantic 
embeddings have used hinge-based triplet ranking 
loss in a cross-modal 
set-up~\cite{UVS,GCSFDIS,andrej,Zhu2015AligningBA}. 
To learn the base embedding network, 
a deep CNN ({\it e.g.}, ResNet152~\cite{resnet}) is 
used as 
an image encoder and a GRU is used as a caption encoder. 
Given an image-caption pair $(I, C)$, the 
triplet loss uses all negative captions 
$\hat{C}$ given $I$, and all negative images $\hat{I}$ 
given $C$, and is given as: 
\begin{eqnarray}
l_{SH}(I,C) = 
\sum_{\hat{C}} [\alpha + s(I,\hat{C}) - s(I,C)]_{+} \nonumber \\+ 
\sum_{\hat{I}}[\alpha + s(\hat{I},C) - s(I,C)]_{+} \;, 
\label{eq:baseEmbed0}
\end{eqnarray}
where $[x]_{+} = max(x,0)$, and $\alpha\ge0$ denotes margin. 
The joint embedding learnt using the above 
{\it sum-of-hinges} loss $l_{SH}$ has been 
popularly termed as baseline 
visual-semantic embedding, or simply VSE. 
Later, \cite{vsepp} showed that the accuracy of 
cross-modal retrieval can be significantly 
improved by considering only the hardest 
negative pairs in 
the above loss. 
For a given anchor point, the hardest negative is the 
negative sample which has the maximum similarity with the 
anchor point. 
Given an image-caption pair $(I, C)$, let 
$I^{*}$ denote 
the hardest negative image for $C$ and let 
$C^{*}$ denote the hardest negative caption for $I$ 
in a mini-batch, and are computed as below: 
\begin{equation}
      I^{*} = \arg\max_{I' \neq I} \; s(z_{I'}, z_{C}), 
\label{eq:hnI} 
\end{equation} 
\begin{equation} 
      C^{*} = \arg\max_{C' \neq C} \; s(z_{I}, z_{C'})  
\label{eq:hnC} 
\end{equation} 
Using these, the new triplet ranking loss is given by: 
\begin{eqnarray}
l_{MH}(I,C) = 
    [\alpha + s(I,C^{*}) - s(I, C)]_{+} \nonumber \\+ 
    [\alpha + s(I^{*},C) - s(I, C)]_{+} \;, 
\label{eq:baseEmbed} 
\end{eqnarray}
The joint embedding learnt using the above 
{\it max-of-hinges} loss $l_{MH}$ was 
termed as visual-semantic embedding with hard 
negatives, or VSE++. Due to the empirical advantages 
of VSE++ over VSE, several subsequent 
approaches~\cite{Engilberge2018FindingBI,Huang2019AnnotationEC,Bastan2020TVSETV,Chen2020LearningJV,Matsubara2021TargetOrientedDO} 
have used VSE++ in their proposals, 
thus making this a de facto baseline for 
visual-semantic embbedings. 
Following these, we also use VSE++ as our 
base joint embedding. 



\section{Our Approach} 
\label{sec:OurApproach}

As we discussed earlier, the ability of triplet 
ranking loss (as in VSE++) to minimize the semantic 
gap 
can be strengthened by reducing the heterogeneity 
gap between incompatible feature distributions 
in diverse modalities, and this can be achieved by 
integrating it with 
contrastive learning. Building upon this hypothesis, 
below we describe our NCE-based 
loss functions for contrastive learning of 
visual-semantic embeddings. In a mini-batch, 
the first loss considers total similarity 
with all the samples from another modality, while 
the second loss considers similarity with only 
the hardest negative 
from another modality. Figure~\ref{fig:illustration} 
gives an overview of both the loss functions. 
As both of our loss functions 
build upon the SimCLR~\cite{simclr} approach, first we 
briefly describe it below. 

\subsection{An overview of SimCLR} 

In SimCLR, a minibatch of $N$ samples is iteratively 
picked from the training 
dataset and two corresponding views are generated for 
each sample. Out of these, one is the original 
sample, and second is obtained by applying 
some transformation(s) to it  
({\it e.g.}, channel distortion, random cropping, etc.), 
thus giving 
$2N$ samples. For each sample, a positive pair is formed 
by considering it along with its transformed view, and 
the remaining $2\times(N-1)$ samples are used to make 
negative pairs. Finally, the similarity (agreement) 
between the 
samples in the positive pair is maximized and the expected 
similarity between the negative pairs is minimized using 
the NCE loss function. 


\subsection{Contrastive Visual-Semantic Embedding} 
\label{sec:CCL} 


Analogous to the unimodal setting of SimCLR, we 
have $2N$ samples 
in a minibatch from $N$ (image,caption) pairs. 
If we directly adapt the SimCLR approach for 
cross-modal data, for each image (caption) we get one 
positive pair and $2\times(N-1)$ negative pairs; 
{\it i.e.}, by pairing a given image/caption 
with the remaining $N-1$ 
images and 
$N-1$ captions, thus giving $2\times(N-1)$ 
negative pairs for each sample. 
This will give us the multi-modal NCE loss that was introduced 
by~\cite{ssmvn} to train a multi-modal versatile 
network (or MVN), which can also be adopted for cross-modal 
matching and retrieval tasks. 
However, as we discussed earlier, in cross-modal retrieval 
tasks, we are interested in reducing the similarity 
between negative pairs across different modalities 
instead of negative pairs within the same modality. 
Hence, for a given image $I$, we propose to 
make the positive pair by 
associating it with its true caption $C$, and 
$N-1$ negative pairs by associating it with the remaining 
$N-1$ captions. 
Similarly, for a given caption $C$, we make the 
positive pair 
by associating it with its true image $I$, and 
$N-1$ negative pairs by associating it with the remaining 
$N-1$ images in the mini-batch. 
Based on this, we define our first contrastive triplet loss 
for 
an image-caption pair $(I, C)$ as: 
\begin{equation}
l_{CSN}(I, C) = 
l_{CSN}^{I}(I, C) + l_{CSN}^{C}(I, C), 
\label{eq:CMCL}
\end{equation} 
\begin{equation} 
l_{CSN}^{I}(I, C) = 
-\text{log}\left(\frac{\text{exp}(s(z_{I}, z_{C}) / \tau )} 
{\Sigma_{k=1}^{N} \text{exp}(s(z_{I}, z_{C_k}) / \tau ) }\right), 
\label{eq:CMCL_I}
\end{equation} 
\begin{equation} 
l_{CSN}^{C}(I, C) = 
- \text{log} \left(\frac{\text{exp}(s(z_{I}, z_{C}) / \tau )} 
{\Sigma_{k=1}^{N} \text{exp}(s(z_{I_k}, z_{C}) / \tau )}\right), 
\label{eq:CMCL_C}
\end{equation} 
where $\tau$ denotes the temperature parameter used 
for scaling the similarity score.
We call the joint embedding learnt using the above 
{\it sum-of-negatives} loss $l_{CSN}$ 
as contrastive visual-semantic 
embedding, or ConVSE. 
As we can observe, the above loss function consists  
of two terms: caption retrieval loss and image retrieval loss, 
with both being given equal importance. 
In Figure~\ref{fig:illustration}(a), 
the left and right portions illustrate the  
terms corresponding to caption and image retrieval losses 
respectively. 
The blue dot denotes the anchor point, green dot 
denotes the positive point and red dots denote 
negative points. 
In the $l_{CSN}$ loss, we aim at pulling the 
positive (target) point closer to the anchor point 
and push all the negatives away from the outer circle 
in a combined manner. 
This is performed 
irrespective of whether the negative points are 
near to the anchor point compared to target or not. 
Considering a case when all the negative points are 
near the outer circle and the target is near the center, 
it is still possible that we get a small positive 
value for the 
$l_{CSN}$ loss 
and update the network, 
which may not be desirable. 
To address this, we next update the $l_{CSN}$ loss 
by considering the influence of only the 
hardest negatives. 

\subsection{Contrastive Visual-Semantic Embedding with Hard Negatives} 
\label{sec:CCLHN} 
The success of cross-modal retrieval tasks 
as measured by Recall@1 (i.e., the first retrieved sample) 
primarily depends on the hard negatives. 
In other words, this would mean that for correct retrieval, we 
want the similarity of the 
positive pair to be the maximum. If the similarity of the 
query (anchor) point with the hardest negative is less 
than the similarity with the target, we can call it 
a successful retrieval. 
Motivated by the empirical gains achieved by the 
usage of hard negatives in VSE++ as discussed in the 
previous section, 
we extend this idea to our contrastive $l_{CSN}$ loss and 
update it by considering the similarity of only the 
hardest negative pair as below: 
\begin{equation}
l_{\widetilde{CMN}}(I, C) = 
l_{\widetilde{CMN}}^{I}(I, C) 
+ l_{\widetilde{CMN}}^{C}(I, C),  
\label{eq:CMCLHN1}
\end{equation} 
\begin{equation} 
l_{\widetilde{CMN}}^{I}(I, C) = 
-\text{log} \left(\frac{\text{exp}(s(z_{I}, z_{C}) / \tau )}
{\text{exp}(s(z_{I}, z_{C^{*}}) / \tau )}\right), 
\end{equation} 
\begin{equation} 
l_{\widetilde{CMN}}^{C}(I, C) = 
-\text{log} \left(\frac{\text{exp}(s(z_{I}, z_{C}) / \tau )}
{\text{exp}(s(z_{I^{*}}, z_{C}) / \tau ) }\right), 
\end{equation}
where, $I^{*}$ (Eq.~\ref{eq:hnI}) 
and $C^{*}$ (Eq.~\ref{eq:hnC}) denote the hardest negative 
image and caption for the anchors $C$ and 
$I$ respectively. 

Now, let us consider the first term 
(i.e., the caption retrieval loss term) of Eq.~\ref{eq:CMCLHN1}. 
Here, we can observe that 
for a positive pair $(I, C)$, if the similarity 
of the image $I$ (anchor) with the target caption $C$ 
is more than the similarity with the hardest negative 
caption $C^{*}$, the loss value would be negative. 
This would subsequently lead to updating the weights of 
the 
network even though the retrieval is correct, which 
is undesirable. 
A similar issue 
will occur when we get a negative value for the 
second term (i.e., the image retrieval loss term). 
To address this, 
we update the loss in Eq.~\ref{eq:CMCLHN1} by 
first taking a max of each term in the loss function with $0$ 
analogous to the conventional hinge-loss, ensuring that 
we do not update the network weights in case of 
correct retrieval. 
Next, we also add a positive margin term $\alpha$ to the 
similarity score of the anchor point with the hardest 
negative before scaling to ensure that 
the hardest negative is less similar than the target by 
a margin. These two changes in $l_{\widetilde{CMN}}$
give us our second contrastive triplet loss as: 
\begin{equation}
l_{CMN}(I, C) = 
l_{CMN}^{I}(I, C) + 
l_{CMN}^{C}(I, C), 
\label{eq:cmnloss} 
\end{equation} 
\begin{equation} 
l_{CMN}^{I}(I, C) = 
\left[-\text{log} \left(\frac{\text{exp}(s(z_{I}, z_{C}) / \tau )}
{\text{exp}((s(z_{I}, z_{C^{*}}) + \alpha) / \tau ) }\right)\right]_{+}, 
\end{equation} 
\begin{equation} 
l_{CMN}^{C}(I, C) = 
\left[-\text{log} \left(\frac{\text{exp}(s(z_{I}, z_{C}) / \tau )} 
{\text{exp}((s(z_{I^{*}}, z_{C}) + \alpha) / \tau ) }\right)\right]_{+}, 
\end{equation} 
We call the joint embedding learnt using the above 
{\it max-of-negatives with margin} loss $l_{CMN}$ 
as contrastive visual-semantic 
embedding with hard negatives, or ConVSE++. 
In Figure~\ref{fig:illustration}(b), we 
illustrate the working of $l_{CMN}$. 
The inner circle represents the maximum desired 
distance (or, the minimum desired similarity) 
of the target sample from the anchor point, and the 
distance between the inner and outer circle 
denotes the margin $\alpha$. 
The hardest negative sample 
(denoted by a continuous red border) 
that lies within the margin 
is used to compute the loss, and the remaining 
negative samples (marked in dashed red 
border) do not contribute in the loss. 
From the figure, we can 
observe that even when all the negatives are within the 
margin, 
the $l_{CMN}$ loss considers only the hardest negative, 
and all the other negatives 
indirectly get pushed away 
by pushing the hardest negative, thus satisfying 
the requirement of cross-modal retrieval tasks where it is 
crucial to obtain correct retrievals among the top few 
retrieved samples. Algorithm 1 summarizes the ConVSE++ method.  

\begin{algorithm}[t] 
  \begin{algorithmic}[1]
    \FOR{sampled minibatch $\{(I_i, C_i)\}_{i=1}^{i=N}$}
      \FOR{\textbf{all} $i \in \{1,\dots,N\}$}
        \STATE $h_{I_i} = ImgEnc(I_i)$ 
        \textcolor{grey}{\hfill\# image representation} 
        \STATE $h_{C_i} = TextEnc(C_i)$ 
        \textcolor{grey}{\hfill\# text representation} 
        \STATE $z_{I_i} = g_I(h_{I_i})$ 
        \textcolor{grey}{\hfill\# projection} 
        \STATE $z_{C_i} = g_C(h_{C_i})$ 
        \textcolor{grey}{\hfill\# projection} 
      \ENDFOR
      
      \FOR{\textbf{all} $i \in \{1,\dots,N\}$ and $j \in \{1,\dots,N\}$}
        \STATE $s(I_i, C_j) = \frac{z_{I_i}^{T} z_{C_j}}{\|z_{I_i}\| \|z_{C_j}\|}$ 
        \textcolor{grey}{\hfill\# pairwise similarity} 
      \ENDFOR
      \FOR{\textbf{all} $i \in \{1,\dots,N\}$}
        \STATE $I_i^{*} = \arg\max_{j \neq i} \; s(z_{I_j}, z_{C_i})$
        \STATE $C_i^{*} = \arg\max_{k \neq i} \; s(z_{I_i}, z_{C_k}) $ 
        \STATE Compute $l_{CMN}(I_i,C_i)$ using Eq.~\ref{eq:cmnloss}. 
      \ENDFOR
      \STATE $\mathcal{L} = \frac{1}{N} \Sigma_{k=1}^{k=N} l_{CMN}(I_k, C_k)$ 
      \STATE Update networks to minimize $\mathcal{L}$
    \ENDFOR
  \end{algorithmic}
   \caption{ConVSE++'s learning algorithm}
\end{algorithm}



\section{Experiments} 
\label{sec:Experiments} 

For experimental analysis, we consider the downstream 
task of cross-modal image-to-text (I2T) and 
text-to-image (T2I) retrieval, and evaluate 
and compare the proposed approaches with 
several existing visual-embedding learning methods. 

\subsection{Experimental Procedure} 
\label{sec:ExperimentalSetup} 

\subsubsection{Datasets and Evaluation metrics} 
We use two datasets in our experiments: 
(1) Flickr30K~\cite{flickr30K} dataset 
which consists of 31000 
images, with each image being described using 
five captions. Out of these, 
we consider 29000, 1000 and 1000 images and corresponding 
captions for training, validation and testing 
respectively following the split from~\cite{andrej}. 
(2) MS-COCO~\cite{mscoco} dataset 
which consists of 113287, 5000 and 5000 images for 
training, validation and testing respectively, with each 
image being described using five captions. 
We use the publicly available splits, and 
report the results after 
averaging over 5 folds of 1k testing images following 
earlier papers.

To evaluate cross-modal retrieval performance, 
we use the metrics used by earlier 
approaches. Precisely, we use 
recall at rank-K (R@K) and report R@1, R@5, and 
R@10 for both I2T and T2I. 
Recall@K measures the percentage of queries for which the 
positive (target) sample from another modality 
is retrieved in the top-K retrieved samples. 
Additionally, we also report ``R@sum'', which 
denotes the summation of all the R@K values. 

\subsubsection{Network architecture}
Our image and text encoders are similar to 
those used in~\cite{vsepp}. Specifically, we use 
ResNet-152~\cite{resnet} pretrained on 
ImageNet as the image encoder. 
Each image is resized to 256 $\times$ 256, and a random 
crop of size 224 $\times$ 224 is given as an input 
to the encoder. This gives a 2048-dimensional 
feature representation 
using the output of the penultimate fully connected 
layer. 
On top of this, we employ a 1024-dimensional fully 
connected layer. 
In the text representation pipeline, we use 
GRU as the caption encoder. 
To the GRU, we feed 300-dimensional word2vec word embeddings 
which produces a 1024-dimensional caption embedding as 
the output. 
We use this architecture as the base embedding network 
to learn a joint embedding space using the $l_{MH}$ loss 
(Eq.~\ref{eq:baseEmbed}). 

For the proposed (ConVSE and ConVSE++) joint embedding 
models, we add a non-linear projection head using two 
MLP layers, on both the image and the text processing pipelines of 
the above base embedding network. 
Both these MLPs consist of two fully connected layers with 
2048 and 1024 hidden units 
respectively, and finally we get 1024-dimensional 
embeddings for each modality. 
This network is trained using the proposed loss functions 
given in Eq.~\ref{eq:CMCL} and Eq.~\ref{eq:cmnloss} to 
obtain ConVSE and ConVSE++ respectively. 
We use the same network architecture for both the datasets. 

\begin{table*}[t] 
\caption{Cross-modal retrieval results of ConVSE and ConVSE++ 
for five runs on the MS-COCO and 
Flickr30K datasets, along with mean and standard 
deviation across all the runs 
} 
\label{tab:CompareRuns} 
\begin{center}
\begin{tabular}{lc|ccccccc|ccccccc} 
\multicolumn{2}{c}{} 
& \multicolumn{7}{c}{\bf MS-COCO}
& \multicolumn{7}{c}{\bf Flickr30K} 
\\ 
\toprule 
\scriptsize{Method} & \scriptsize{Run} 
& \multicolumn{3}{c}{\scriptsize{Image-to-Text (I2T)}} 
& \multicolumn{3}{c}{\scriptsize{Text-to-Image (T2I)}} 
& 
& \multicolumn{3}{c}{\scriptsize{Image-to-Text (I2T)}} 
& \multicolumn{3}{c}{\scriptsize{Text-to-Image (T2I)}} 
& 
\\ 
& 
& \scriptsize{R@1} & \scriptsize{R@5} & \scriptsize{R@10} 
& \scriptsize{R@1} & \scriptsize{R@5} & \scriptsize{R@10} 
& \scriptsize{R@sum} 
& \scriptsize{R@1} & \scriptsize{R@5} & \scriptsize{R@10} 
& \scriptsize{R@1} & \scriptsize{R@5} & \scriptsize{R@10} 
& \scriptsize{R@sum} 
\\ 
\midrule 
\multirow{7}{*}{\scriptsize{ConVSE}} & \scriptsize{1} 
& \scriptsize{63.2} & \scriptsize{89.2} & \scriptsize{95.2} 
& \scriptsize{50.3} & \scriptsize{83.5} & \scriptsize{92.2} 
& \scriptsize{473.6} 
& \scriptsize{54.2} & \scriptsize{79.6} & \scriptsize{88.1} 
& \scriptsize{39.6} & \scriptsize{70.3} & \scriptsize{80.4} 
& \scriptsize{412.2} 
\\ 
& \scriptsize{2} 
& \scriptsize{63.4} & \scriptsize{89.2} & \scriptsize{95.2} 
& \scriptsize{50.4} & \scriptsize{83.5} & \scriptsize{92.2} 
& \scriptsize{473.9} 
& \scriptsize{54.9} & \scriptsize{79.3} & \scriptsize{88.6} 
& \scriptsize{39.4} & \scriptsize{70.1} & \scriptsize{80.4} 
& \scriptsize{412.7} 
\\ 
& \scriptsize{3} 
& \scriptsize{63.9} & \scriptsize{89.3} & \scriptsize{95.3} 
& \scriptsize{50.5} & \scriptsize{83.3} & \scriptsize{92.3} 
& \scriptsize{474.6} 
& \scriptsize{55.6} & \scriptsize{81.0} & \scriptsize{88.8} 
& \scriptsize{39.0} & \scriptsize{69.3} & \scriptsize{79.4} 
& \scriptsize{413.1} 
\\ 
& \scriptsize{4} 
& \scriptsize{63.6} & \scriptsize{89.3} & \scriptsize{95.4} 
& \scriptsize{50.7} & \scriptsize{83.4} & \scriptsize{92.3} 
& \scriptsize{474.7} 
& \scriptsize{56.4} & \scriptsize{80.0} & \scriptsize{87.9} 
& \scriptsize{39.8} & \scriptsize{70.3} & \scriptsize{80.2} 
& \scriptsize{414.6} 
\\ 
& \scriptsize{5} 
& \scriptsize{63.8} & \scriptsize{89.5} & \scriptsize{95.4} 
& \scriptsize{50.7} & \scriptsize{83.7} & \scriptsize{92.2} 
& \scriptsize{475.3} 
& \scriptsize{56.1} & \scriptsize{80.4} & \scriptsize{88.2} 
& \scriptsize{39.4} & \scriptsize{70.7} & \scriptsize{80.1} 
& \scriptsize{414.9} 
\\ 
\cmidrule{2-16}
& \scriptsize{mean} 
& \scriptsize{63.6} & \scriptsize{89.3} & \scriptsize{95.3} 
& \scriptsize{50.5} & \scriptsize{83.5} & \scriptsize{92.2} 
& \scriptsize{474.4} 
& \scriptsize{55.4} & \scriptsize{80.1} & \scriptsize{88.3} 
& \scriptsize{39.4} & \scriptsize{70.1} & \scriptsize{80.1} 
& \scriptsize{413.5} 
\\ 
& \scriptsize{(std)} 
& \scriptsize{(0.29)} & \scriptsize{(0.12)} & \scriptsize{(0.10)} 
& \scriptsize{(0.18)} & \scriptsize{(0.15)} & \scriptsize{(0.05)} 
& \scriptsize{(0.68)} 
& \scriptsize{(0.90)} & \scriptsize{(0.67)} & \scriptsize{(0.37)} 
& \scriptsize{(0.30)} & \scriptsize{(0.52)} & \scriptsize{(0.41)} 
& \scriptsize{(1.19)} 
\\ 
\midrule 
\multirow{7}{*}{\scriptsize{ConVSE++}} & \scriptsize{1} 
& \scriptsize{66.7} & \scriptsize{91.1} & \scriptsize{96.7} 
& \scriptsize{53.0} & \scriptsize{84.1} & \scriptsize{91.7} 
& \scriptsize{483.3} 
& \scriptsize{58.2} & \scriptsize{82.3} & \scriptsize{89.8} 
& \scriptsize{41.4} & \scriptsize{70.5} & \scriptsize{79.5} 
& \scriptsize{421.7} 
\\ 
& \scriptsize{2} 
& \scriptsize{66.6} & \scriptsize{91.4} & \scriptsize{96.8} 
& \scriptsize{53.0} & \scriptsize{84.1} & \scriptsize{91.4} 
& \scriptsize{483.3} 
& \scriptsize{57.8} & \scriptsize{83.6} & \scriptsize{89.6} 
& \scriptsize{41.0} & \scriptsize{70.5} & \scriptsize{79.8} 
& \scriptsize{422.3} 
\\ 
& \scriptsize{3} 
& \scriptsize{66.6} & \scriptsize{91.3} & \scriptsize{96.7} 
& \scriptsize{53.0} & \scriptsize{84.1} & \scriptsize{91.8} 
& \scriptsize{483.5} 
& \scriptsize{57.8} & \scriptsize{84.3} & \scriptsize{88.9} 
& \scriptsize{41.0} & \scriptsize{71.1} & \scriptsize{80.0} 
& \scriptsize{423.1} 
\\ 
& \scriptsize{4} 
& \scriptsize{66.7} & \scriptsize{91.5} & \scriptsize{96.6} 
& \scriptsize{53.1} & \scriptsize{84.1} & \scriptsize{91.6} 
& \scriptsize{483.6} 
& \scriptsize{57.9} & \scriptsize{83.5} & \scriptsize{90.6} 
& \scriptsize{41.3} & \scriptsize{70.6} & \scriptsize{79.4} 
& \scriptsize{423.3} 
\\ 
& \scriptsize{5} 
& \scriptsize{66.7} & \scriptsize{91.5} & \scriptsize{96.7} 
& \scriptsize{53.1} & \scriptsize{84.1} & \scriptsize{91.5} 
& \scriptsize{483.6} 
& \scriptsize{58.9} & \scriptsize{83.3} & \scriptsize{89.8} 
& \scriptsize{41.8} & \scriptsize{71.1} & \scriptsize{79.7} 
& \scriptsize{424.6} 
\\ 
\cmidrule{2-16}
& \scriptsize{mean} 
& \scriptsize{66.7} & \scriptsize{91.4} & \scriptsize{96.7} 
& \scriptsize{53.0} & \scriptsize{84.1} & \scriptsize{91.6} 
& \scriptsize{483.5} 
& \scriptsize{58.1} & \scriptsize{83.4} & \scriptsize{89.7} 
& \scriptsize{41.3} & \scriptsize{70.8} & \scriptsize{79.7} 
& \scriptsize{423.0} 
\\ 
& \scriptsize{(std)} 
& \scriptsize{(0.05)} & \scriptsize{(0.17)} & \scriptsize{(0.07)} 
& \scriptsize{(0.05)} & \scriptsize{(0.00)} & \scriptsize{(0.16)} 
& \scriptsize{(0.15)} 
& \scriptsize{(0.47)} & \scriptsize{(0.72)} & \scriptsize{(0.61)} 
& \scriptsize{(0.33)} & \scriptsize{(0.31)} & \scriptsize{(0.24)} 
& \scriptsize{(1.10)} 
\\ 
\bottomrule 
\end{tabular}
\end{center}
\end{table*}

\begin{table}[t] 
\caption{Cross-modal retrieval results of ConVSE++ 
for different values of temperature ($\tau$) 
and projection output dimensionality ($d$) 
on the MS-COCO dataset 
} 
\label{tab:VaryHyperparameters} 
\begin{center}
\begin{tabular}{lcccccccc}
\toprule 
\scriptsize{Hyp.} 
& \scriptsize{Val.} 
& \multicolumn{3}{c}{\scriptsize{Image-to-Text (I2T)}} 
& \multicolumn{3}{c}{\scriptsize{Text-to-Image (T2I)}} 
& \\ 
& 
& \scriptsize{R@1} & \scriptsize{R@5} & \scriptsize{R@10} 
& \scriptsize{R@1} & \scriptsize{R@5} & \scriptsize{R@10} 
& \scriptsize{R@sum} 
\\ 
\midrule 
\multirow{4}{*}{\scriptsize{$\tau$}} 
& \scriptsize{0.05} 
& \scriptsize{66.0} & \scriptsize{91.2} & \scriptsize{96.3} 
& \scriptsize{52.9} & \scriptsize{84.0} & \scriptsize{91.5} 
& \scriptsize{481.9} 
\\ 
& \scriptsize{\bf 0.1} 
& \scriptsize{\bf 66.6} & \scriptsize{\bf 91.3} & \scriptsize{\bf 96.7} 
& \scriptsize{\bf 53.0} & \scriptsize{\bf 84.1} & \scriptsize{\bf 91.8} 
& \scriptsize{\bf 483.5} 
\\ 
& \scriptsize{0.5} 
& \scriptsize{66.5} & \scriptsize{90.8} & \scriptsize{96.8} 
& \scriptsize{52.9} & \scriptsize{84.2} & \scriptsize{91.5} 
& \scriptsize{482.7} 
\\ 
& \scriptsize{1.0} 
& \scriptsize{66.3} & \scriptsize{90.9} & \scriptsize{96.6} 
& \scriptsize{52.8} & \scriptsize{84.0} & \scriptsize{91.4} 
& \scriptsize{482.0} 
\\ 
\midrule 
\multirow{6}{*}{\scriptsize{$d$}} 
& \scriptsize{64} 
& \scriptsize{64.5} & \scriptsize{90.1} & \scriptsize{96.0} 
& \scriptsize{51.3} & \scriptsize{82.8} & \scriptsize{90.7} 
& \scriptsize{475.4} 
\\ 
& \scriptsize{128} 
& \scriptsize{65.8} & \scriptsize{90.7} & \scriptsize{96.2} 
& \scriptsize{52.4} & \scriptsize{83.9} & \scriptsize{91.2} 
& \scriptsize{480.2} 
\\ 
& \scriptsize{256} 
& \scriptsize{66.2} & \scriptsize{91.1} & \scriptsize{96.6} 
& \scriptsize{52.7} & \scriptsize{83.9} & \scriptsize{91.5} 
& \scriptsize{482.0} 
\\ 
& \scriptsize{512} 
& \scriptsize{66.1} & \scriptsize{91.1} & \scriptsize{96.6} 
& \scriptsize{52.6} & \scriptsize{84.0} & \scriptsize{91.5} 
& \scriptsize{481.9} 
\\ 
& \scriptsize{\bf 1024} 
& \scriptsize{\bf 66.6} & \scriptsize{\bf 91.3} & \scriptsize{\bf 96.7} 
& \scriptsize{\bf 53.0} & \scriptsize{\bf 84.1} & \scriptsize{\bf 91.8} 
& \scriptsize{\bf 483.5} 
\\ 
& \scriptsize{2048} 
& \scriptsize{66.5} & \scriptsize{91.2} & \scriptsize{96.8} 
& \scriptsize{52.8} & \scriptsize{84.0} & \scriptsize{91.5} 
& \scriptsize{482.8} 
\\ 
\bottomrule 
\end{tabular}
\end{center}
\end{table}

\subsubsection{Training details} 

We train the base embedding network by keeping the image encoder fixed for 30 epochs with a learning rate of 0.0002 for the first 15 epochs and then update the learning rate to 0.00002 for the subsequent epochs. After this, we fine-tune the base embedding network, including the image encoder, with a learning rate of 0.00002 for 15 epochs.
The minibatch size is set to 128 during the initial training and fine-tuning of the base embedding network.
For the training of our ConVSE and 
ConVSE++ joint embedding networks, we initialize the 
network weights with the previously 
trained base embedding network, and 
then train the whole network for 30 epochs by keeping a fixed learning rate of 0.00002.
The weights of the image encoder remain freezed during this training. 
We conducted multiple experiments for training the 
ConVSE and ConVSE++ embedding networks 
by setting the batch size to 256, and 
varying 
size of the common embedding space in \{64, 128, 256, 512, 1024, 2048\}, 
and the temperature parameter ($\tau$) in \{0.05, 0.1, 0.5, 1.0\}, and 
found the optimal values of the common embedding and 
temperature to be 1024 and 0.1 respectively. 
In all the experiments, we fixed the margin value to $\alpha=0.2$, used 
Adam~\cite{adam} for optimization. 
For both the datasets, we use the same training set-up. 
Our implementation and pretrained models can be downloaded from 
this \href{https://drive.google.com/file/d/183Ov69JyYPyz742qVeR1IjJV3QyB20Xw/view?usp=sharing}{link}. 

\begin{table*}[t] 
\caption{Cross-modal retrieval comparisons of the 
proposed joint embedding ConVSE based on sum-of-negatives 
contrastive loss 
(Eq.~\ref{eq:CMCL}) and ConVSE++ based on max-of-negatives 
contrastive loss 
(Eq.~\ref{eq:cmnloss}), with three 
baseline joint embedding methods: an adaptation of the 
multi-modal contrastive loss of the MVN~\cite{ssmvn} approach 
for visual-semantic embedding, 
VSE that uses sum-of-hinges triplet ranking loss 
(Eq.~\ref{eq:baseEmbed0}), and VSE++ that uses 
max-of-hinges triplet ranking loss (Eq.~\ref{eq:baseEmbed}) 
} 
\label{tab:BaselineComparison} 
\begin{center}
\begin{tabular}{ll|ccccccc|ccccccc} 
\multicolumn{2}{c}{} 
& \multicolumn{7}{c}{\scriptsize{\bf MS-COCO}} 
& \multicolumn{7}{c}{\scriptsize{\bf Flickr30K}} 
\\
\toprule 
\scriptsize{Method} & \scriptsize{Backbone} 
& \multicolumn{3}{c}{\scriptsize{Image-to-Text (I2T)}} 
& \multicolumn{3}{c}{\scriptsize{Text-to-Image (T2I)}} 
& 
& \multicolumn{3}{c}{\scriptsize{Image-to-Text (I2T)}} 
& \multicolumn{3}{c}{\scriptsize{Text-to-Image (T2I)}} 
& 
\\ 
& \scriptsize{Network} 
& \scriptsize{R@1} & \scriptsize{R@5} & \scriptsize{R@10} 
& \scriptsize{R@1} & \scriptsize{R@5} & \scriptsize{R@10} 
& \scriptsize{R@sum} 
& \scriptsize{R@1} & \scriptsize{R@5} & \scriptsize{R@10} 
& \scriptsize{R@1} & \scriptsize{R@5} & \scriptsize{R@10} 
& \scriptsize{R@sum} 
\\ 
\midrule 
\scriptsize{MVN~\cite{ssmvn}} 
& \scriptsize{ResNet-152} 
& \scriptsize{54.1} & \scriptsize{83.8} & \scriptsize{92.1} 
& \scriptsize{44.9} & \scriptsize{80.5} & \scriptsize{90.5} 
& \scriptsize{445.9} 
& \scriptsize{42.8} & \scriptsize{72.9} & \scriptsize{83.1} 
& \scriptsize{34.1} & \scriptsize{64.7} & \scriptsize{75.6} 
& \scriptsize{373.2} 
\\ 
\scriptsize{VSE} 
& \scriptsize{ResNet-152} 
& \scriptsize{56.0}	& \scriptsize{85.8}	& \scriptsize{93.5}	
& \scriptsize{43.7}	& \scriptsize{79.4}	& \scriptsize{89.7}	
& \scriptsize{448.1}
& \scriptsize{42.1}	& \scriptsize{73.2}	& \scriptsize{84.0}	
& \scriptsize{31.8}	& \scriptsize{62.6}	& \scriptsize{74.1}	
& \scriptsize{367.8}
\\ 
\scriptsize{VSE++} 
& \scriptsize{ResNet-152} 
& \scriptsize{64.6}	& \scriptsize{90.0}	& \scriptsize{{95.7}}	
& \scriptsize{{52.0}}	& \scriptsize{{\bf 84.3}} & \scriptsize{{92.0}}	
& \scriptsize{{478.6}}
& \scriptsize{52.9}  	& \scriptsize{80.5}  	& \scriptsize{87.2}  	
& \scriptsize{39.6}  	& \scriptsize{70.1}  	& \scriptsize{79.5}  	
& \scriptsize{409.8}
\\ 
\scriptsize{ConVSE (Ours)} 
& \scriptsize{ResNet-152} 
& \scriptsize{63.9} & \scriptsize{89.3} & \scriptsize{95.3} 
& \scriptsize{50.5} & \scriptsize{83.3} & \scriptsize{\bf 92.3} 
& \scriptsize{474.6} 
& \scriptsize{55.6} & \scriptsize{81.0} & \scriptsize{88.8} 
& \scriptsize{39.0} & \scriptsize{69.3} & \scriptsize{79.4} 
& \scriptsize{413.1} 
\\ 
\scriptsize{ConVSE++ (Ours)} 
& \scriptsize{ResNet-152} 
& \scriptsize{\bf 66.7} & \scriptsize{\bf 91.5} & \scriptsize{\bf 96.6} 
& \scriptsize{\bf 53.1} & \scriptsize{84.1} & \scriptsize{91.6} 
& \scriptsize{\bf 483.6} 
& \scriptsize{\bf 57.8} & \scriptsize{\bf 84.3} & \scriptsize{\bf 88.9} 
& \scriptsize{\bf 41.0} & \scriptsize{\bf 71.1} & \scriptsize{\bf 80.0} 
& \scriptsize{\bf 423.1} 
\\ 
\bottomrule 
\end{tabular}
\end{center}
\end{table*}

\begin{table*}[t] 
\caption{Cross-modal retrieval results using the proposed 
approaches and state-of-the-art methods on the MS-COCO and 
Flickr30K datasets. 
The best results in both are highlighted in bold} 
\label{tab:ComparisonMscoco} 
\begin{center}
\begin{tabular}{ll|ccccccc|ccccccc}
\multicolumn{2}{c}{} 
& \multicolumn{7}{c}{\scriptsize{\bf MS-COCO}} 
& \multicolumn{7}{c}{\scriptsize{\bf Flickr30K}} 
\\
\toprule 
\scriptsize{Method} & \scriptsize{Backbone} 
& \multicolumn{3}{c}{\scriptsize{Image-to-Text (I2T)}} 
& \multicolumn{3}{c}{\scriptsize{Text-to-Image (T2I)}} 
& 
& \multicolumn{3}{c}{\scriptsize{Image-to-Text (I2T)}} 
& \multicolumn{3}{c}{\scriptsize{Text-to-Image (T2I)}} 
& 
\\ 
& \scriptsize{Network} 
& \scriptsize{R@1} & \scriptsize{R@5} & \scriptsize{R@10} 
& \scriptsize{R@1} & \scriptsize{R@5} & \scriptsize{R@10} 
& \scriptsize{R@sum} 
& \scriptsize{R@1} & \scriptsize{R@5} & \scriptsize{R@10} 
& \scriptsize{R@1} & \scriptsize{R@5} & \scriptsize{R@10} 
& \scriptsize{R@sum} 
\\ 
\midrule 
\scriptsize{UVS~\cite{UVS}} 
& \scriptsize{AlexNet} 
& \scriptsize{23.0}	& \scriptsize{50.7}	& \scriptsize{62.9}	
& \scriptsize{16.8}	& \scriptsize{42.0}	& \scriptsize{56.5}	
& \scriptsize{251.9}  
& \scriptsize{23.0}  & \scriptsize{50.7}  & \scriptsize{62.9}  
& \scriptsize{16.8}  & \scriptsize{42.0}  & \scriptsize{56.5}  
& \scriptsize{251.9}  
\\ 
\scriptsize{m-RNN~\cite{mRNN}} 
& \scriptsize{VGG} 
& \scriptsize{43.4}  	& \scriptsize{75.7}  	& \scriptsize{85.8}  	
& \scriptsize{31.0}  	& \scriptsize{66.7}  	& \scriptsize{79.9}  	
& \scriptsize{382.5}  
& \scriptsize{35.4}  & \scriptsize{63.8}    & \scriptsize{73.7}   
& \scriptsize{22.8}    & \scriptsize{50.7}    & \scriptsize{63.1}    
& \scriptsize{309.5}  
\\ 
\scriptsize{DSPE+FV~\cite{DSPEFV}} 
& \scriptsize{VGG} 
& \scriptsize{50.1}	& \scriptsize{79.7}	& \scriptsize{89.2}	
& \scriptsize{39.6}	& \scriptsize{75.2}	& \scriptsize{86.9}	
& \scriptsize{420.7} 
& \scriptsize{40.3}  & \scriptsize{68.9}  & \scriptsize{79.9}  
& \scriptsize{29.7}  & \scriptsize{60.1}  & \scriptsize{72.1}  
& \scriptsize{351.0}
\\ 
\scriptsize{sm-LSTM~\cite{smLSTM}} 
& \scriptsize{VGG} 
& \scriptsize{53.2}	& \scriptsize{83.1}	& \scriptsize{91.5}	
& \scriptsize{40.7}	& \scriptsize{75.8}	& \scriptsize{87.4}	
& \scriptsize{431.7} 
& \scriptsize{42.5}  & \scriptsize{71.9}  & \scriptsize{81.5}  
& \scriptsize{30.2}  & \scriptsize{60.4}  & \scriptsize{72.3}  
& \scriptsize{358.8}  
\\
\scriptsize{RRF-Net~\cite{RRFNet}} 
& \scriptsize{ResNet-152} 
& \scriptsize{56.4}	& \scriptsize{85.3}	& \scriptsize{91.5}	
& \scriptsize{43.9}	& \scriptsize{78.1}	& \scriptsize{88.6}	
& \scriptsize{443.8}
& \scriptsize{47.6}  & \scriptsize{77.4}  & \scriptsize{87.1}  
& \scriptsize{35.4}  & \scriptsize{68.3}  & \scriptsize{79.9}  
& \scriptsize{395.7}
\\ 
\scriptsize{DAN~\cite{DAN}} 
& \scriptsize{ResNet-152} 
& \scriptsize{-} & \scriptsize{-} & \scriptsize{-} 
& \scriptsize{-} & \scriptsize{-} & \scriptsize{-} 
& \scriptsize{-} 
& \scriptsize{55.0}  & \scriptsize{81.8}  & \scriptsize{89.0}  
& \scriptsize{39.4}  & \scriptsize{69.2}  & \scriptsize{79.1}  
& \scriptsize{413.5} 
\\
\scriptsize{Embedding Net~\cite{embeddingNet}} 
& \scriptsize{VGG} 
& \scriptsize{50.4}	& \scriptsize{79.3}	& \scriptsize{69.4}	
& \scriptsize{39.8}   & \scriptsize{75.3}   & \scriptsize{86.6}   
& \scriptsize{400.8} 
& \scriptsize{40.7}  & \scriptsize{69.7}  & \scriptsize{79.2}  
& \scriptsize{29.2}   & \scriptsize{59.6}   & \scriptsize{71.7}   
& \scriptsize{350.1}
\\ 
\scriptsize{CMPM+CMPC~\cite{CMPMCMPC}} 
& \scriptsize{MobileNet} 
& \scriptsize{52.9}	& \scriptsize{83.8}	& \scriptsize{92.1}	
& \scriptsize{41.3}	& \scriptsize{74.6}	& \scriptsize{85.9}	
& \scriptsize{430.6}
& \scriptsize{40.3}  & \scriptsize{66.9}  & \scriptsize{76.7}  
& \scriptsize{30.4}  & \scriptsize{58.2}  & \scriptsize{68.5}  
& \scriptsize{341.0}
\\ 
\scriptsize{CMPM+CMPC~\cite{CMPMCMPC}} 
& \scriptsize{ResNet-152} 
& \scriptsize{-} & \scriptsize{-} & \scriptsize{-} 
& \scriptsize{-} & \scriptsize{-} & \scriptsize{-} 
& \scriptsize{-} 
& \scriptsize{49.6}  & \scriptsize{76.8}  & \scriptsize{86.1}  
& \scriptsize{37.3}  & \scriptsize{65.7}  & \scriptsize{75.5}  
& \scriptsize{391.0}
\\ 
\scriptsize{Joint Learning~\cite{jointLearning}} 
& \scriptsize{ResNet-152} 
& \scriptsize{55.3}	& \scriptsize{82.7}	& \scriptsize{90.2}	
& \scriptsize{41.7}	& \scriptsize{75.0}	& \scriptsize{87.4}	
& \scriptsize{432.3} 
& \scriptsize{48.6}  & \scriptsize{73.6}  & \scriptsize{83.6}  
& \scriptsize{32.3}  & \scriptsize{62.5}  & \scriptsize{74.0}  
& \scriptsize{374.6}
\\ 
\scriptsize{TIMAM~\cite{TIMAM}} 
& \scriptsize{ResNet-152} 
& \scriptsize{-} & \scriptsize{-} & \scriptsize{-} 
& \scriptsize{-} & \scriptsize{-} & \scriptsize{-} 
& \scriptsize{-} 
& \scriptsize{53.1}  & \scriptsize{78.8}  & \scriptsize{87.6}  
& \scriptsize{42.6}  & \scriptsize{71.6}  & \scriptsize{81.9}  
& \scriptsize{415.6}
\\ 
\scriptsize{Dual-path I~\cite{dualPathStage}} 
& \scriptsize{ResNet-152} 
& \scriptsize{52.2}	& \scriptsize{80.4}	& \scriptsize{88.7}	
& \scriptsize{37.2}	& \scriptsize{69.5}	& \scriptsize{80.6}	
& \scriptsize{408.6}
& \scriptsize{44.2}  & \scriptsize{70.2}  & \scriptsize{79.7}  
& \scriptsize{30.7}  & \scriptsize{59.2}  & \scriptsize{70.8}  
& \scriptsize{354.8}
\\ 
\scriptsize{Dual-path II~\cite{dualPathStage}} 
& \scriptsize{ResNet-152} 
& \scriptsize{\textbf{65.6}}	& \scriptsize{\textbf{89.8}}	& \scriptsize{\bf 95.5}	
& \scriptsize{47.1}	& \scriptsize{79.9}	& \scriptsize{90.0}	
& \scriptsize{\bf 467.9}
& \scriptsize{55.6}  & \scriptsize{81.9}  & \scriptsize{89.5}  
& \scriptsize{39.1}  & \scriptsize{69.2}  & \scriptsize{\textbf{80.9}} 
& \scriptsize{416.2}
\\ 
\scriptsize{ITMeetsAL~\cite{ITMeetsAL}} 
& \scriptsize{MobileNet} 
& \scriptsize{54.7}	& \scriptsize{84.3}	& \scriptsize{91.1}	
& \scriptsize{41.0}	& \scriptsize{76.7}	& \scriptsize{88.1}	
& \scriptsize{435.9}
& \scriptsize{46.6}  & \scriptsize{73.5}  & \scriptsize{82.5}  
& \scriptsize{34.4}  & \scriptsize{63.3}  & \scriptsize{74.2}  
& \scriptsize{374.5}
\\ 
\scriptsize{ITMeetsAL~\cite{ITMeetsAL}} 
& \scriptsize{ResNet-152} 
& \scriptsize{58.5}	& \scriptsize{85.3}	& \scriptsize{92.1}	
& \scriptsize{\bf 48.3}	& \scriptsize{\bf 82.0}	& \scriptsize{\bf 90.6}	
& \scriptsize{456.8}
& \scriptsize{\textbf{56.5}} & \scriptsize{\textbf{82.2}} & \scriptsize{\textbf{89.6}} 
& \scriptsize{\textbf{43.5}} & \scriptsize{\textbf{71.8}} & \scriptsize{80.2}  
& \scriptsize{\textbf{423.8}}
\\ 
\midrule 
\scriptsize{ConVSE (Ours)} 
& \scriptsize{ResNet-152} 
& \scriptsize{63.9} & \scriptsize{89.3} & \scriptsize{95.3} 
& \scriptsize{50.5} & \scriptsize{83.3} & \scriptsize{\bf 92.3} 
& \scriptsize{474.6} 
& \scriptsize{55.6} & \scriptsize{81.0} & \scriptsize{88.8} 
& \scriptsize{39.0} & \scriptsize{69.3} & \scriptsize{79.4} 
& \scriptsize{413.1}
\\ 
\scriptsize{ConVSE++ (Ours)} 
& \scriptsize{ResNet-152} 
& \scriptsize{\bf 66.7} & \scriptsize{\bf 91.5} & \scriptsize{\bf 96.6} 
& \scriptsize{\bf 53.1} & \scriptsize{\bf 84.1} & \scriptsize{91.6} 
& \scriptsize{\bf 483.6} 
& \scriptsize{\bf 57.8} & \scriptsize{\bf 84.3} & \scriptsize{\bf 88.9} 
& \scriptsize{\bf 41.0} & \scriptsize{\bf 71.1} & \scriptsize{\bf 80.0} 
& \scriptsize{\bf 423.1} 
\\ 
\bottomrule 
\end{tabular}
\end{center}
\end{table*}

\subsection{Ablation studies} 

In Table~\ref{tab:CompareRuns}, we show the results of 
the proposed ConVSE and ConVSE++ approaches for five 
runs on the MS-COCO dataset. Here, we can observe that 
their performance is 
quite stable, with standard deviation below $0.50\%$ 
in almost all the cases. For comparisons with existing 
methods, we use the median results with respect to 
R@sum. 
In Table~\ref{tab:VaryHyperparameters}, we show the 
performance of ConVSE++ for different values of 
the hyperparameters temperature ($\tau$) 
and projection output 
dimensionality, keeping other hyperparameters as fixed. 
Here, we can observe that the 
performance is quite stable with respect to temperature. 
In case of projection output dimensionality, initially 
the performance is somewhat low for smaller dimensionality 
values, however it increases quickly and then becomes 
stable. It is worth noting that similar trend is 
observed with respect to these two hyperparameters in 
SimCLR~\cite{simclr} as well, though for unimodal data.

In general, the above analyses show that the performance of 
our proposed approaches are quite stable with respect to 
multiple runs and variations in the values of different 
hyperparameters, thus making them easily reproducible.

\subsection{Results and Discussion} 
\label{sec:ResultsAndDiscussion} 


\begin{figure*}[h] 
\centering
\begin{tabular}{|p{0.30\linewidth}|p{0.30\linewidth}|p{0.30\linewidth}|}
\toprule 
  \includegraphics[width=5.35cm,height=3cm]{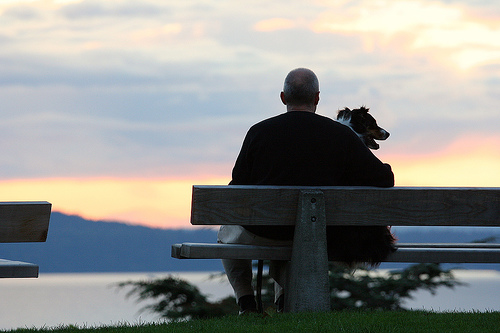}
& \includegraphics[width=5.35cm,height=3cm]{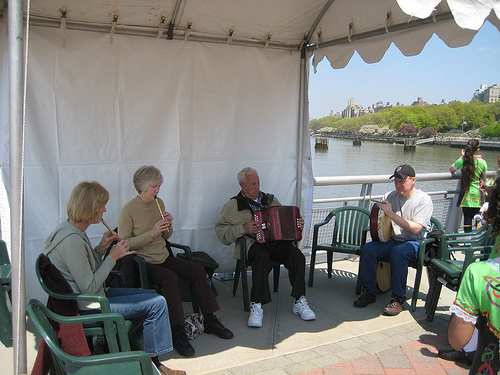}
& \includegraphics[width=5.35cm,height=3cm]{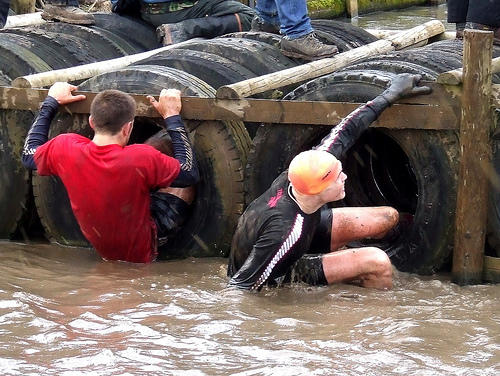} 
\\ 
\textcolor{mygreen}{\bf 
(1) A man holding a dog sitting on a bench overlooking a lake.
$\;\;\;\;\;\;\;\;\;\;\;\;\;\;$
(2) A man and his dog watch the sunset from a bench.
$\;\;\;\;\;\;\;\;\;\;\;\;\;\;\;\;\;\;\;\;\;$
(3) A man and a dog sit on a bench near a body of water. 
} 
& 
\textcolor{mygreen}{\bf 
(1) Four elderly people sitting under a white tent playing musical instruments. 
$\;\;\;\;\;\;\;\;\;\;\;\;\;\;\;\;\;\;\;\;\;\;\;\;\;\;$
(2) Four people playing instruments underneath a white tent. 
$\;\;\;\;\;\;\;\;\;\;\;\;\;\;\;\;\;\;\;\;\;\;\;\;\;\;\;\;\;\;\;\;\;\;\;\;\;$
(3) A group of people sit outside under a white shade tent, sitting in green lawn chairs, playing instruments with a view of a body of water behind them. 
} 
& 
\textcolor{mygreen}{\bf 
(1) One man in the hole of a tire in water, and another man getting up from the water next to him.
$\;\;\;\;\;\;\;\;\;\;\;\;\;\;\;\;\;\;\;\;\;\;\;\;\;\;\;\;\;\;\;\;\;\;\;$
(2) Two men are playing with tires in muddy water.
$\;\;\;\;\;\;\;\;\;\;\;\;\;\;\;\;\;\;\;\;\;\;\;\;\;\;\;\;\;\;\;\;\;\;\;\;\;$
(3) Two men climb onto tires while sitting in the muddy water. 
} 
\\ 
\midrule 
  \includegraphics[width=5.35cm,height=3cm]{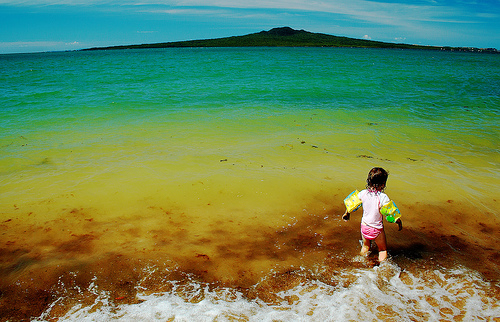}
& \includegraphics[width=5.35cm,height=3cm]{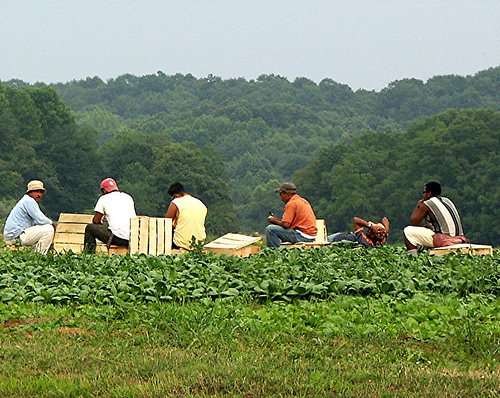}
& \includegraphics[width=5.35cm,height=3cm]{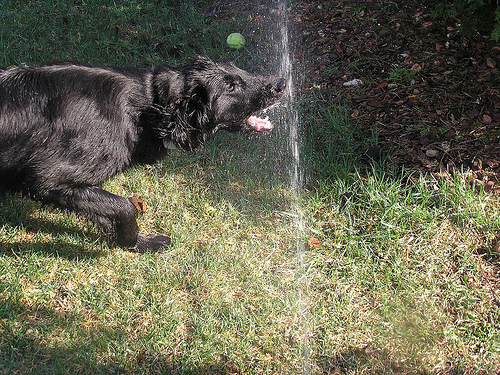} 
\\ 
\textcolor{red}{\bf 
(1) A young girl is running across the beach to get to the water. 
$\;\;\;\;\;\;\;\;\;\;\;\;\;\;\;\;\;\;\;\;\;\;\;\;\;\;\;\;\;\;\;\;\;\;\;\;\;$
(2) A little girl wearing pink is running along a beach towards the water. 
$\;\;\;\;\;\;\;\;\;\;\;\;\;\;\;\;\;\;\;\;\;\;\;\;\;\;\;\;\;\;\;\;\;\;\;\;\;\;\;\;\;\;\;\;$
} 
\textcolor{mygreen}{\bf 
(3) A girl at the shore of a beach with a mountain in the distance. 
} 
& \textcolor{red}{\bf 
(1) Four people relaxing on a grassy hill overlooking a rocky valley. 
$\;\;\;\;\;\;\;\;\;\;\;\;\;\;\;\;\;\;\;\;\;\;\;\;\;\;\;\;\;\;\;\;\;\;\;\;\;\;\;\;\;\;\;\;$
(2) An African group of men, women, and babies pose in a field with large hills in the background. 
$\;\;\;\;\;\;\;\;\;\;\;\;\;\;\;\;\;\;\;\;\;\;\;\;\;\;\;\;\;\;\;\;\;\;\;\;\;\;\;\;\;\;\;\;$
(3) A group of people rest on a cliff overlooking a scenic view. 
} 
&  
\textcolor{red}{\bf 
(1) A black dog springs up into a pool. 
$\;\;\;\;\;\;\;\;\;\;\;\;\;\;\;\;\;\;\;\;\;\;\;\;\;\;\;\;\;\;\;\;\;\;\;\;\;\;\;\;\;\;\;\;$
(2) A small black dog jumping over gates. 
$\;\;\;\;\;\;\;\;\;\;\;\;\;\;\;\;\;\;\;\;\;\;\;\;\;\;\;\;\;\;\;\;\;\;\;\;\;\;\;\;\;\;\;\;$
(3) A black dog sprints near the blue fence. 
} 
\\ 
\bottomrule 
\end{tabular}
\caption{Examples of image-to-text (I2T) 
    retrieval on the Flickr30K dataset using the 
    using the proposed ConVSE++ approach. For each query image, we 
    show the top-3 retrieved captions. 
    The captions in green denote correct retrievals, and 
    those in red denote incorrect retrievals. 
    Top: success cases; 
    bottom: failure cases. 
    } 
\label{fig:ExamplesI2T} 
\end{figure*}

\subsubsection{Comparison with baselines} 

In Table~\ref{tab:BaselineComparison}, we compare the 
proposed joint embedding ConVSE and ConVSE++ with 
with the baseline joint embedding methods VSE, VSE++ and 
an adaptation of the 
multi-modal contrastive loss of~\cite{ssmvn} 
for learning a joint image-caption embedding space. We can 
make following observations from these results: 
(a) In general, our best method ConVSE++ achieves best results. 
Specifically, it achieves absolute improvements of 
5.0 and 10.0 respectively in R@sum on the MS-COCO and 
Flickr30K datasets respectively compared to the second best method. 
(b) The results obtained using MVN 
are comparable to VSE and consistently inferior to ConVSE, 
thus confirming the advantage of excluding the modality of the 
anchor point while defining negative pairs. 
(c) ConVSE always outperforms VSE, and ConVSE++ always outperforms 
VSE++. These results validate our initial hypothesis that 
the ability of triplet ranking loss functions in learning 
joint embedding spaces can be strengthened by integrating them 
with contrastive training, as it helps in reducing the 
heterogeneity gap between incompatible feature distributions of 
diverse modalities. 
(d) In general, ConVSE++ outperforms ConVSE, and 
VSE++ outperforms VSE. 
The improvements are particularly significant 
w.r.t. R@1 and R@5 in comparison to R@10 
for both I2T and T2I and on both the datasets. 
These results validate that 
using only the hardest negative instead of all 
can be particularly useful in learning visual-semantic 
embeddings for cross-modal retrieval tasks, where it is 
desirable to have high 
accuracy among the top few retrievals.

\subsubsection{Comparison with benchmark methods} 

In Table~\ref{tab:ComparisonMscoco}, 
we compare the results of cross-modal retrieval using 
the proposed approaches 
with the state-of-the-art visual-semantic embedding 
methods on the MS-COCO and Flickr30K datasets 
respectively. 
Note that in almost all the recent approaches, the 
backbone vision network is the same ({\it i.e.}, ResNet-152). 
For both I2T and T2I 
tasks on the MS-COCO dataset, both ConVSE and 
ConVSE++ outperform all the compared 
methods, achieving absolute improvements of 
6.7 and 15.7 respectively in R@sum. 
Similarly, on the Flickr30K dataset, our ConVSE++ approach 
outperforms all the compared approaches except the 
recent ITMeetsAL~\cite{ITMeetsAL} approach where it 
achieves comparable results. 
These results demonstrate 
the empirical promise of using ConVSE and ConVSE++ 
for cross-modal retrieval compared to the existing approaches.

\subsubsection{Qualitative Analyses} 
\label{sec:QualitativeAnalysis} 

\begin{figure*}[h] 
\centering
\begin{tabular}{|p{0.30\linewidth}p{0.30\linewidth}p{0.30\linewidth}|}
\toprule 
\multicolumn{3}{|c|}{\bf Query: Chinese male on his cellphone walking through the alley.} 
\\ 
  \fcolorbox{mygreen}{mygreen}{\includegraphics[width=5.15cm,height=3cm]{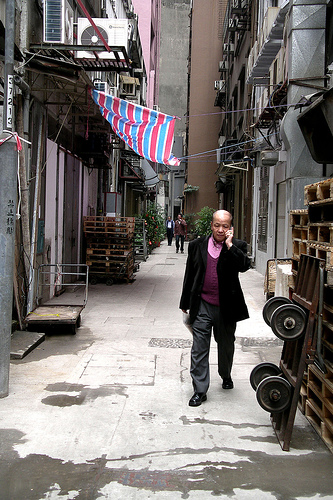}} 
& \fcolorbox{red}{red}{\includegraphics[width=5.15cm,height=3cm]{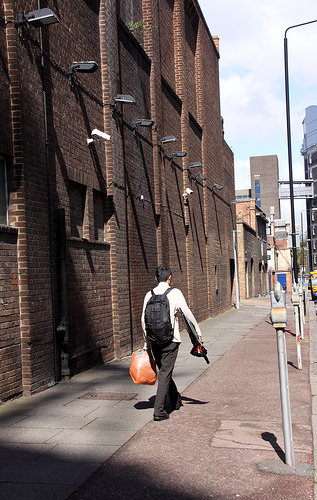}} 
& \fcolorbox{red}{red}{\includegraphics[width=5.15cm,height=3cm]{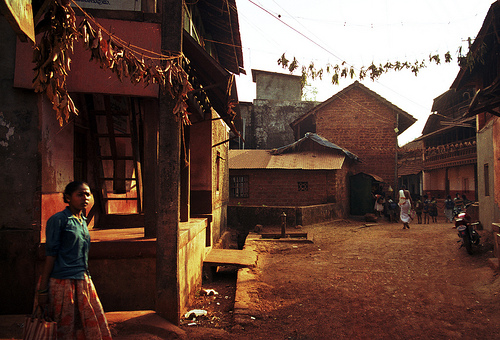}} 
\\ 
\midrule 
\multicolumn{3}{|c|}{\bf Query: Three brown dogs are jumping up at the woman wearing blue.} 
\\ 
  \fcolorbox{mygreen}{mygreen}{\includegraphics[width=5.15cm,height=3cm]{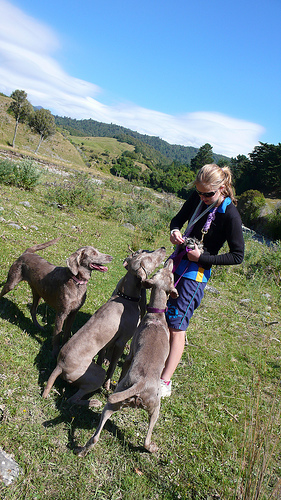}} 
& \fcolorbox{red}{red}{\includegraphics[width=5.15cm,height=3cm]{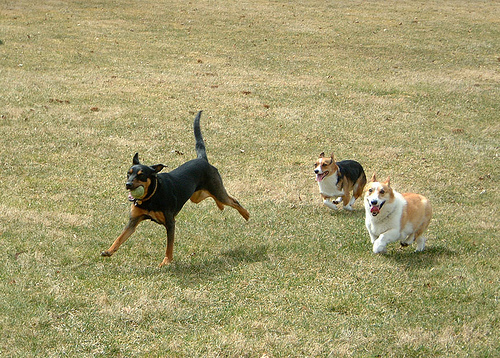}} 
& \fcolorbox{red}{red}{\includegraphics[width=5.15cm,height=3cm]{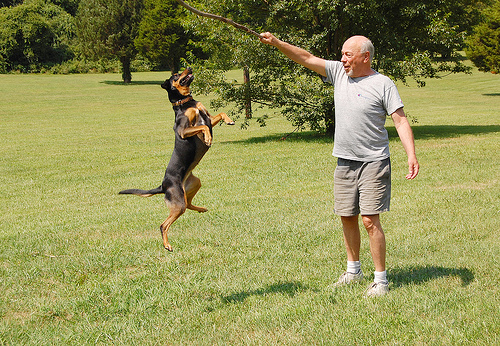}} 
\\ 
\bottomrule 
\multicolumn{3}{c}{}  
\\ 
\toprule 
\multicolumn{3}{|c|}{\bf Query: A child is splashing in the water.} 
\\ 
  \fcolorbox{red}{red}{\includegraphics[width=5.15cm,height=3cm]{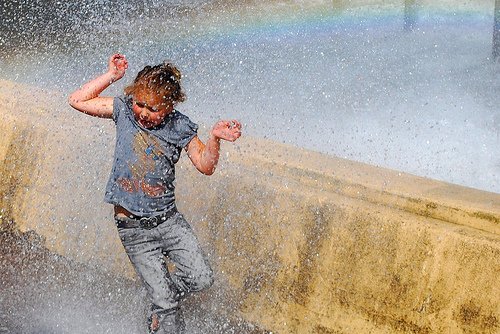}} 
& \fcolorbox{red}{red}{\includegraphics[width=5.15cm,height=3cm]{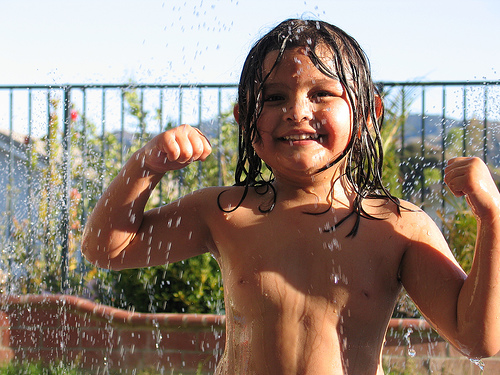}} 
& \fcolorbox{red}{red}{\includegraphics[width=5.15cm,height=3cm]{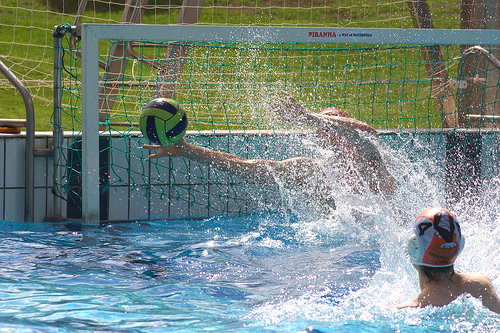}} 
\\ 
\midrule 
\multicolumn{3}{|c|}{\bf Query: A woman with a black shirt and tan apron is standing behind a counter in a restaurant.} 
\\ 
  \fcolorbox{red}{red}{\includegraphics[width=5.15cm,height=3cm]{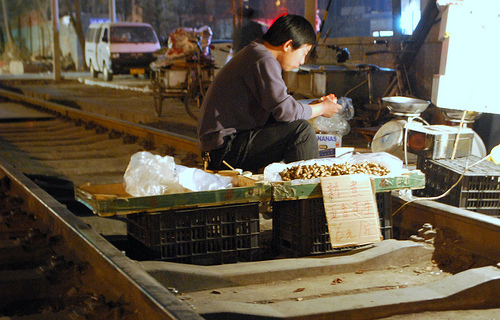}} 
& \fcolorbox{red}{red}{\includegraphics[width=5.15cm,height=3cm]{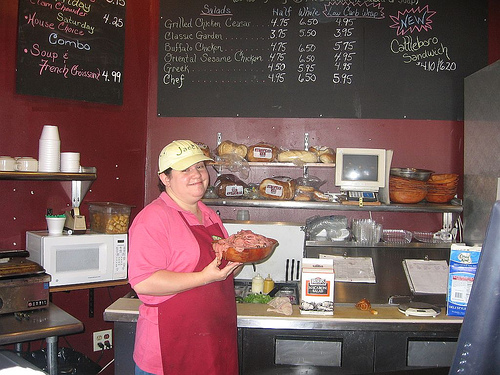}} 
& \fcolorbox{mygreen}{mygreen}{\includegraphics[width=5.15cm,height=3cm]{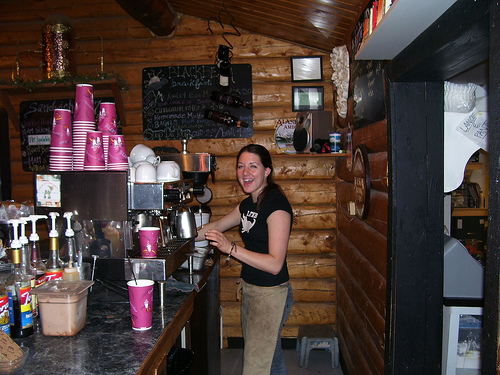}} 
\\ 
\bottomrule 
\end{tabular}
\caption{Examples of text-to-image (T2I) 
    retrieval on the Flickr30K dataset using the 
    using the proposed ConVSE++ approach. For each query 
    caption, we show the top-3 retrieved images 
    (from left to right). 
    The images with green border denote correct retrievals, and 
    those with red border denote incorrect retrievals. 
    The top two rows denote examples of success cases where 
    the correct image is retrieved at rank-1, and 
    the bottom two rows denote examples of failure cases where the 
    correct image is not retrieved at rank-1. 
    } 
\label{fig:ExamplesT2I} 
\end{figure*}

In Figure~\ref{fig:ExamplesI2T} and~\ref{fig:ExamplesT2I}, we 
show some examples of image-to-text (I2T) and text-to-image (T2I) 
retrieval respectively. For both these tasks, we show examples of 
both success and failure cases. 
Particularly in case of incorrect retrievals, we can observe that 
they are semantically quite related to the corresponding query. 
In general, both the success as well as failure cases validate 
the capability of our approach in learning complex 
visual-semantic relationships.

\section{Conclusion} 
\label{sec:Conclusion} 

Triplet loss and contrastive training are two powerful paradigms 
for learning joint visual-semantic embeddings. 
In this paper, we have proposed two new formulations 
that integrate them in a principled way, and subsequently 
achieve results either better than or comparable to the 
state-of-the-art methods on the downstream task of 
cross-modal image-text retrieval. 
Looking at the generality of our formulations and 
their performance trends, we believe they 
can also be adopted for other cross-modal 
learning and matching tasks.

\section*{Acknowledgment}

YV would like to thank the Department of Science and
Technology (India) for the INSPIRE Faculty Award 2017. 

\bibliographystyle{IEEEtran} 
\bibliography{egbib}

\end{document}